\documentclass[final]{cvpr}

\newcommand{\final}{0}

\usepackage{times}
\usepackage{epsfig}
\usepackage{graphicx}
\usepackage{amsmath}
\usepackage{amssymb}
\usepackage{amsthm}
\usepackage{subcaption}
\usepackage{url}
\usepackage{gensymb}
\usepackage{ulem}
\usepackage[symbol]{footmisc}
\usepackage{makecell}
\usepackage{xcolor}

\ifdefined\siggraph
\usepackage{times}
\fi

\usepackage{color}
\usepackage{ifthen}
\usepackage{float}
\usepackage{alltt}
\usepackage{newlfont} 
\usepackage{array}

\usepackage{wrapfig}
\usepackage{booktabs}
\usepackage{multirow}

\newcommand{\PreserveBackslash}[1]{\let\temp=\\#1\let\\=\temp}

\definecolor{DeltaColor}{rgb}{0.039,0.73,0.71}
\definecolor{SetaColor}{rgb}{0.867, 0.0235, 0.376}
\definecolor{SigmaColor}{rgb}{0.98,0.45,0.0}
\definecolor{RedColor}{rgb}{0.8,0,0}
\definecolor{AlphaColor}{rgb}{0,0,0.8}
\definecolor{BetaColor}{rgb}{0.8,0,0.8}
\definecolor{GammaColor}{rgb}{0.5,0,0.7}
\definecolor{EpsilonColor}{rgb}{0.353,0.725,0.906}
\definecolor{TauColor}{rgb}{0.423,0.235,0.192}
\newcommand{\weikai}[1]{{\color{RedColor} Weikai: #1 $\qed$}}
\newcommand{\haiwei}[1]{{\color{AlphaColor} Haiwei: #1 $\qed$}}
\newcommand{\shichen}[1]{{\color{GammaColor} Shichen: #1 $\qed$}}
\newcommand{\hao}[1]{{\color{HaoColor} Hao: #1 $\qed$}}

\newcommand{\Note}[1]{{\it\color{blue} #1}}
\newcommand{\nothing}[1]{}

\definecolor{AudioColor}{rgb}{0.56,0.34,0.62}

\definecolor{DeadlineColor}{rgb}{0.9,0.4,0} 

\definecolor{figred}{rgb}{1,0,0}
\definecolor{figgreen}{rgb}{0,0.6,0}
\definecolor{figblue}{rgb}{0,0,1}
\definecolor{figpink}{rgb}{1,0.63,0.63}

\ifthenelse{\equal{\final}{1}}
{
	\renewcommand{\weikai}[1]{}
	\renewcommand{\haiwei}[1]{}
	\renewcommand{\shichen}[1]{}
	\renewcommand{\hao}[1]{}
	\renewcommand{\Note}[1]{}
}
{}

\newcounter{pccount}
\floatstyle{plain}

\newcommand{\filename}[1]{\url{#1}}
\newcommand{\foldername}[1]{\url{#1}}

\newcommand{\convNameFull}{SE(3) separable point convolution}
\newcommand{\ConvNameFull}{SE(3) separable point convolution}
\newcommand{\convNameShort}{SPConv}

\newcommand{\convNameIntra}{SE(3) group convolution}
\newcommand{\ConvNameIntra}{SE(3) group convolution}
\newcommand{\convNameInter}{SE(3) point convolution}
\newcommand{\ConvNameInter}{SE(3) point convolution}

\newcommand{\poolingNameShort}{GA pooling}
\newcommand{\poolingName}{group attentive pooling}

\newcommand{\soThree}{\text{SO(3)}}
\newcommand{\seThree}{\text{SE(3)}}

\newcommand{\point}{x}
\newcommand{\pointi}{x_i}
\newcommand{\Point}{\mathcal{P}}

\newcommand{\SoFeat}{\mathcal{F}}

\newcommand{\soAnc}{g}
\newcommand{\soAncj}{g_j}

\newcommand{\SoAncGroup}{G}

\newcommand{\kernelFunc}{h}
\newcommand{\kernelFuncInter}{h_1}
\newcommand{\kernelFuncIntra}{h_2}
\newcommand{\interCorrelation}{\kappa}
\newcommand{\kernelSize}{K}
\newcommand{\kernelWeight}{W}

\newcommand{\SphFeat}{\mathcal{F}}

\newcommand{\kernelNum}{K}

\newcommand{\rotation}{\mathcal{R}}

\usepackage[pagebackref=true,breaklinks=true,colorlinks,bookmarks=false]{hyperref}

\def\cvprPaperID{6371} 
\def\confYear{CVPR 2021}

\graphicspath{
	{figs/raster/}
	{figs/handdrawn/}
}

\begin{document}
\title{Equivariant Point Network for 3D Point Cloud Analysis}

%

	\author{Haiwei Chen$^{1,2}$, Shichen Liu$^{1,2}$, Weikai Chen$^{2,3}$, Hao Li$^{4}$ \\
		$^1$University of Southern California\\
		$^2$USC Institute for Creative Technologies\\
		$^3$Tencent Game AI Research Center\\
		$^4$Pinscreen\\
		{\tt\small \{chenh,lshichen,hill\}@ict.usc.edu \tt chenwk891@gmail.com \tt hao@hao-li.com }
	}

\maketitle


\begin{abstract}
	
Features that are equivariant to a larger group of symmetries have been shown to be more discriminative and powerful in recent studies~\cite{cohen2016group,weiler2018learning,cohen2018spherical}. 
However, higher-order equivariant features often come with an exponentially-growing computational cost. 
Furthermore, it remains relatively less explored how rotation-equivariant features can be leveraged to tackle 3D shape alignment tasks. While many past approaches have been based on either non-equivariant or invariant descriptors to align 3D shapes, we argue that such tasks may benefit greatly from an equivariant framework. 
%
In this paper, we propose an effective and practical \seThree{} (3D translation and rotation) equivariant network for point cloud analysis that addresses both problems. 
First, we present \convNameFull{}, a novel framework that breaks down the 6D convolution into two separable convolutional operators alternatively performed in the 3D Euclidean and \soThree{} spaces respectively. This significantly reduces the computational cost without compromising the performance. 
Second, we introduce an attention layer to effectively harness the expressiveness of the equivariant features. While jointly trained with the network, the attention layer implicitly derives the intrinsic local frame in the feature space and generates attention vectors that can be integrated with different alignment tasks.
We evaluate our approach through extensive studies and visual interpretations. The empirical results demonstrate that our proposed model outperforms strong baselines in a variety of benchmarks. \textit{Code is available at \href{https://github.com/nintendops/EPN_PointCloud}{https://github.com/nintendops/EPN\_PointCloud}.}
%
%

\end{abstract}

\section{Introduction}
\label{sec:intro}

\begin{figure}[htb]
 \begin{center}
  \centering
  \includegraphics[width=1.0\linewidth]{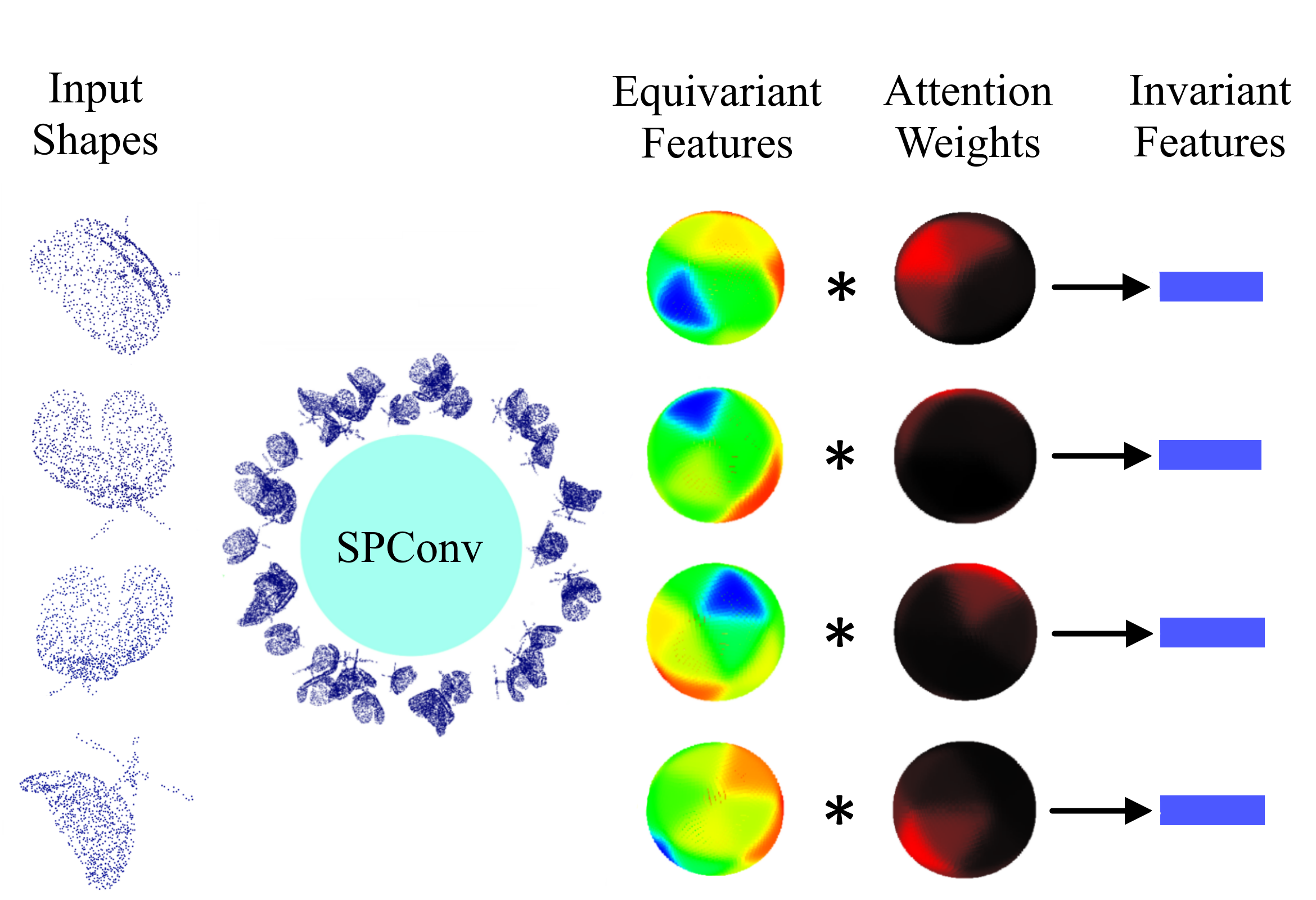}
 \end{center}
 \vspace{-16pt}
 \caption{ The core of our network is a convolution operator on point clouds, termed \ConvNameFull{} (\convNameShort{}), that consumes features defined in the \seThree{} space and outputs per-point features that are \seThree{} equivariant. 
 When the output feature is spatially pooled over the Euclidean space, it becomes \soThree{} equivariant, as visualized above by projecting onto the spherical domain.
 Our method also supports a faithful conversion from the equivariant feature to its invariant counterpart by using a novel attentive fusing mechanism.
Thereby, we offer a general framework that can generate equivariant or invariant point feature depending on the nature of downstream applications.
}
 
 \label{fig:teaser}
 \vspace{-15pt}
\end{figure}

The success of 2D CNNs stems in large part from the ability of exploiting the translational symmetries via weight sharing and translation equivariance.
Recent trends strive to duplicate this success to 3D domain in order to shed new light on the 3D learning tasks.
With the 3D scanning technology being the mainstream manner of measuring the real world, point cloud arises naturally as one of the most prominent 3D representations.
Yet, despite its simple and unified structure, it remains a nuisance to extend the CNN architecture to analyzing point clouds. 
In addition, the group of transformations in 3D data is more complex compared to 2D images, as 3D entities are often transformed by arbitrary 3D translations and 3D rotations when observed. 
Although group-invariant operators could render identical features even under different group transforms, it fails to distinguish distinct instances with internal symmetries (e.g. the counterparts of ``6" and ``9" in 3D scenarios regarding rotational symmetry). In contrast, equivariant features are much more expressive thanks to their ability to retain information about the input group transform on the feature maps throughout the neural layers.
As a result, it could be very beneficial for point cloud features to be equivariant to the \seThree{} group of transformations while being invariant to point permutations.

Despite the importance of deriving \seThree{}-equivariant features for point clouds, progress in this regard remains highly sparse. The main obstacles arise in two aspects.
First, the cost of computing convolutions between 6-dimensional functions over the entire \seThree{} spaces is prohibitive especially in the presence of bulky 3D raw scans.
Second, it remains challenging to fully harness the expressiveness of equivariant features without losing important structural information at a low computational cost. 
In particular, matching any two group-equivariant features is the prerequisite of many applications like correspondence computation, pose estimation, etc. One common practice is to compute the best relative group transformation that maximizes the similarity of the input features when the transformation is applied. This typically requires solving a PnP optimization which is quite costly considering the high dimensionality of the features. Another option is to fuse the equivariant features into invariant ones via pooling operation and directly compare the invariant features to obtain similarity. However, we argue that the naive pooling operations will inevitably discard useful features and damage the equivariant structure of the feature. 

In this paper, we strive to address both of the problems by introducing an effective and practical framework for learning \seThree{}-equivariant features of point clouds.
In particular, inspired by the spirit of ``going wider" in the Inception module~\cite{szegedy2015going}, we first propose \textit{\seThree{} separable convolution}, a novel paradigm that breaks down the naive 6D convolution into two separable convolutional operators alternatively performed in the 3D Euclidean and \soThree{} spaces. Due to the non-commutative and non-compact nature of \seThree{} group, it is non-trivial to factorize \seThree{} convolution into two separable sub-operators. We achieve this goal by first lifting the input points to the homogeneous space. We then take advantage of the finite rotation groups such as the icosahedral and aggregate spatially-convoluted features as functions on the rotation groups that are processed via group convolution.  
The proposed \seThree{} separable convolution significantly reduces the computational cost of a \seThree{} convolution and leads to practical solutions that can be deployed in the commodity hardware.

Second, we present an attention mechanism specially tailored for fusing \seThree{}-equivariant features. We observe that while the commonly used pooling operations, such as max or mean pooling, work well in translation equivariant networks like 2D CNNs, they are not best suited for fusing equivariant features in \soThree{} groups. This is mostly due to the highly sparse and non-linear structure of \soThree{} features which poses additional challenges for max/mean pooling to maintain its unique pattern without losing too much information.
We introduce \textit{\poolingName{}} (\poolingNameShort{}) to adaptively fuse rotation-equivariant features into their invariant counterparts. 
Trained together with the network, the \poolingNameShort{} layer implicitly learns an intrinsic local frame of the feature space and generates attention weights to guide the pooling of rotation-equivariant features.

Third, compared to invariant features, equivariant features preserves, rather than discards, spatial structure and therefore can be seen as a more discriminative representation. It is for this reason that translational equivariance has been the premise for convolutional approaches for detection and instance segmentation~\cite{girshick2014rich}. Similarly, through the attention mechanism, the equivariant framework can be utilized for inferring 3D rotations. We demonstrate in the experiments that this structure significantly outperforms a non-equivariant framework in a shape alignment task.

We validate our proposed framework on a variety of tasks. 
Experimental results show that our approach consistently outperforms strong baselines. We also perform ablation analysis and qualitative visualization to evaluate the effectiveness of each algorithmic component.

\vspace{-3mm}
\section{Related Work}
\label{sec:related_work}


\paragraph{Learning-based Point Descriptor.}

The seminal work on handling irregular structure of point cloud places the main emphasis on permutation-invariant functions~\cite{qi2017pointnet}. Later works proposes shift equivariant hierarchical architectures with localized filters to align with the regular grid CNNs~\cite{qi2017pointnet++,li2018so,liu2019point2sequence}. Explicit convolution kernels have also received tremendous attention in recent years. In particular, various kernel forms have been proposed, including voxel bins~\cite{hua2018pointwise}, polynomial functions~\cite{xu2018spidercnn} or linear functions~\cite{groh2018flex}. Other works consider different representations of point clouds, noticeably image projection~\cite{elbaz20173d,huang2018learning,li2020end} and voxels~\cite{maturana2015voxnet,ben20183dmfv,qi2016volumetric,wu20153d}. We point interested readers to~\cite{guo2020deep} for a comprehensive survey on point cloud convolution.

Rotation invariant point descriptors have been an active research area due to its importance to correspondence matching. While the features extracted by most of the above approaches are permutation-invariant, very few of them can achieve rotation-invariance. The Perfect Match~\cite{gojcic2019perfect} incorporates a local reference frame (LRF) to extract rotation-invariant features from the voxelized point cloud. Similarly,~\cite{zhao2019quaternion} proposes a capsule network that consumes a point cloud along with the estimated LRF to disentangle shape and pose information. By only taking point pair as input, PPF-FoldNet~\cite{deng2018ppf} can learn rotation-invariant descriptors using folding-based encoding of point pair features. However, invariant features may be limited in expressiveness as spatial information is discarded a priori.

\vspace{-4mm}
\paragraph{Learning Rotation-equivariant Features.}

Since CNNs are sensitive to rotations, a rapidly growing body of work focus on investigating rotation-equivariant variants. Starting from the 2D domain, various approaches have been proposed to achieve rotation equivariance by applying multiple oriented filters~\cite{diego2017}, performing a log-polar transform of the input~\cite{esteves2017polar}, replacing filters with circular harmonics~\cite{worrall2017harmonic} or rotating the filters~\cite{li2018deep,weiler2018learning}. Cohen and Welling later extend the domain of 2D CNNs from translation to finite groups~\cite{cohen2016group} and further to arbitrary compact groups~\cite{cohen2016steerable}.

When it comes to the domain of 3D rotation, previous efforts can be divided into spectral and non-spectral methods. In the spectral branch, generalized Fourier transform for $S^2$ and $\soThree$ underlies designs for rotation equivariant CNN. We would like to highlight two seminal works~\cite{cohen2018spherical},~\cite{esteves2018learning} that define convolution operators respectively by spherical ($S^2$) correlation, and $\soThree{}$ correlation with circularly symmetric kernels. The works most relevant to our setting are extensions of the two spectral paradigms to the 3D spatial domain. A number of works extend spherical CNNs to 3D voxels grids~\cite{worrall2018cubenet,weiler20183d,esteves2018learning,jiang2019spherical}. Yet, the research work on exploring the potential on point clouds remains sparse, with the exception of a concurrent work Tensor field network (TFN)~\cite{thomas2018tensor}, which achieves \seThree{} equivariance on irregular point clouds. While~\cite{thomas2018tensor} shares with us in the use of tensor-field representation, their proposed filters are products of radial function and spherical harmonics. We instead focus on a non-spectral, computationally efficient separable framework.

Our work finds inspiration from the non-spectral group equivariant approaches that have seen recent progress, extending from mathematical framework derived in~\cite{cohen2016group, cohen2019gauge}. Specifically,~\cite{cohen2019gauge} provides a general framework for the practical implementation of convolution on discretized rotation group, with icosahedral convolution as an examplar. Discrete group convolution characterizes many recent works on images~\cite{esteves2019equivariant, lenssen2018group}, spherical signal~\cite{spezialetti2019learning}, voxel grid~\cite{worrall2018cubenet} and point cloud~\cite{li2019discrete}. Most of these works focus only on rotational equivariance. We are the first in this branch to provide a unified, hierarchical framework for point cloud convolution that is equivariant to the space of $\seThree$.

\section{Method}
\label{sec:method}

\begin{figure*}[t]
 \begin{center}
  \centering
  \includegraphics[width=0.9\linewidth]{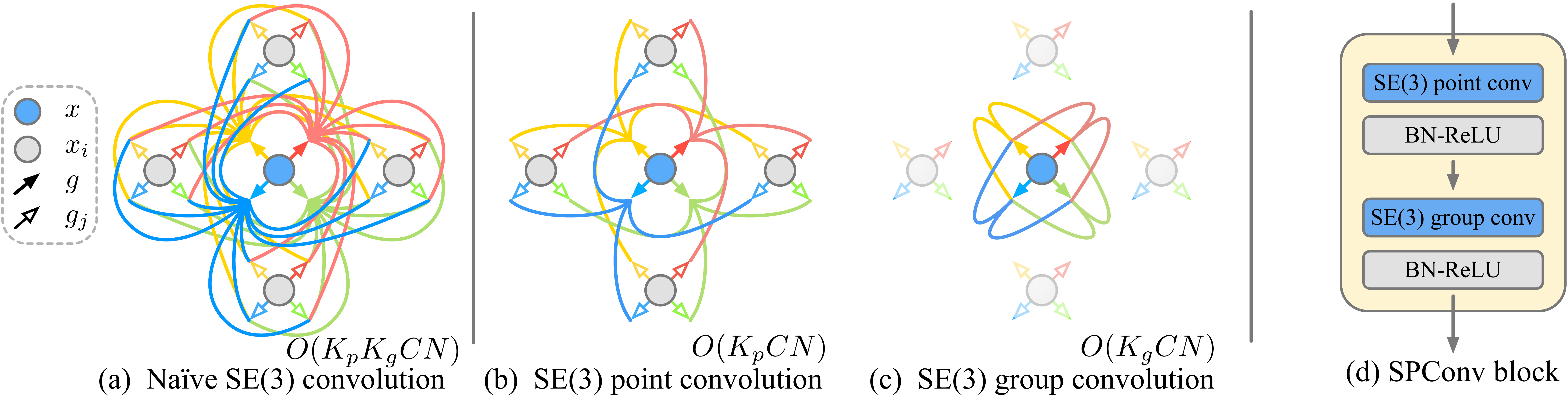}
 \end{center}
  \vspace{-5mm}
 \caption{Illustration of \convNameShort{}. Each arrow represents an element in the group and each edge represents a correlation needed to compute in the convolution operator. We propose to use two separable convolutions (b)(c) to achieve \seThree{} equivariance. The computational cost is much lower than the naive 6D convolution (a). (d) shows the structure of a basic SPConv block.}{}

 \label{fig:conv}
 \vspace{-5mm}
\end{figure*}

\paragraph{Overview.}

In this section, we first start with the preliminaries of \seThree{} convolutions. We will then provide the detailed mathematical formulation of our approach: (1) the \seThree{} separable convolution; and (2) attention mechanism for the equivariant features. 
The Lie group \seThree{} is the group of rigid body transformations in three-dimensions: 
\begin{align*}
\seThree = \{ A \vert A = 
\begin{bmatrix}
	R & t \\
	0 & 1
\end{bmatrix}, R \in  \soThree, t \in  \mathbb{R}^{3}\}. 
\end{align*}

\noindent \seThree{} is homeomorphic to $\mathbb{R}^{3} \times \soThree$. Therefore, a function that is equivariant to \seThree{} must be equivariant to both 3D translation $ t \in \mathbb{R}^{3}$ and 3D rotation $ g \in \soThree $. Given a spatial point $\point$ and a rotation $\soAnc$, let us first define the continuous feature representation in \seThree{} as a function $\SoFeat(\pointi,\soAncj): \mathbb{R}^3 \times \soThree \rightarrow \mathbb{R}^D$. Equivariance to \seThree{} is expressed as satisfying $ \forall A \in \seThree,
  A(\SphFeat \ast \kernelFunc)(\point, \soAnc) = (A\SphFeat \ast \kernelFunc)(\point, \soAnc) 
$. The \seThree{} equivariant continuous convolutional operator can be defined as 
\vspace{-2mm}
\begin{multline}
  (\SoFeat \ast \kernelFunc)(\point, \soAnc) \\ =
  \int_{\pointi \in \mathbb{R}^{3}} \int_{\soAncj \in \soThree} \SoFeat(\pointi, \soAncj) \kernelFunc(\soAnc^{-1}(\point - \pointi), \soAncj^{-1}\soAnc)
  \label{equ:6d_conv_continuous},
\end{multline}

\noindent where $\kernelFunc$ is a kernel $\kernelFunc(\point,\soAnc): \mathbb{R}^3 \times \soThree \rightarrow \mathbb{R}^D$. The convolution is computed by translating and rotating the kernel and then computing a dot product with the input function $\SoFeat$. 
We prove that this convolution is equivariant to \seThree{} in the supplementals.

\vspace{-4mm}
\paragraph{Discretization.} To discretize Eq.~\ref{equ:6d_conv_continuous}, we starts with discretizing the \seThree{} space into a composition of a finite set of 3D spatial point $\Point: \{ \point \vert \point \in  \mathbb{R}^{3} \}$ and a finite rotation group $\SoAncGroup \subset \soThree$.  
 This leads to a discrete \seThree{} feature mapping function $\SoFeat(\pointi,\soAncj): \Point \times \SoAncGroup \rightarrow \mathbb{R}^D$. The discrete convolutional operator in \seThree{} is therefore:

\begin{equation}
  (\SoFeat \ast \kernelFunc)(\point, \soAnc) = \sum_{\pointi \in \Point} \sum_{\soAncj \in \SoAncGroup} \SoFeat(\pointi, \soAncj) \kernelFunc(\soAnc^{-1} (\point - \pointi), \soAncj^{-1}\soAnc).
  \label{equ:6d_conv}
\end{equation}

\noindent We note that such discretization serves as a good approximation of the continuous formulation in Eq.~\ref{equ:6d_conv_continuous}, where the approximation error can be further mitigated by the rotation augmentation~\cite{azulay2018deep}. If we interpret $\Point$ as a set of 3D displacements, this leads to an equivalent definition:
\begin{equation}
\begin{split}
  (\SoFeat \ast \kernelFunc)(\point, \soAnc) &= \sum_{\pointi \in \Point} \sum_{\soAncj \in \SoAncGroup} \SoFeat(\soAnc^{-1}(\point - \pointi), \soAncj^{-1}\soAnc) \kernelFunc(\pointi, \soAncj) \\
  &=  \sum_{\pointi^\prime \in \Point_{\soAnc}} \sum_{\soAncj \in \SoAncGroup} \SoFeat(\point - \pointi^\prime, \soAncj^{-1}\soAnc) \kernelFunc(\soAnc\pointi^\prime, \soAncj).
  \end{split}
  \label{equ:6d_conv_alt}
\end{equation}

\vspace{-1mm}
Without loss of generality, we assume the coordinate is expressed in the local frame of $\point$ and therefore $g^{-1}x = x$. In the second row of Eq.~\ref{equ:6d_conv_alt}, the summation over the set $\Point$ becomes a summation over a rotated set $\Point_\soAnc: \{ \soAnc^{-1}\point \vert \point \in \Point \}$. When written this way, we can see that the kernel is parameterized by a set of translation offsets and rotation offsets under the reference frame given by $\soAnc$.  We call the discrete set $\Point \times \SoAncGroup$ the domain of the kernel with a kernel size of $\vert \Point \vert \times \vert \SoAncGroup \vert$.
\vspace{-3mm}
\paragraph{\seThree{} Separable Convolution.}  A key issue with Eq.~\ref{equ:6d_conv_alt} is that the convolution is computed over a 6-dimensional space -- a naive implementation would be computational prohibitive. Inspired by the core idea of separable convolution~\cite{chollet2017xception}, we observe that the kernel $\kernelFunc$ with a kernel size $\vert \Point \vert \times \vert \SoAncGroup \vert$ can be separated into two smaller kernels, denoting $\kernelFuncInter$ with a kernel size of $\vert \Point \vert \times 1 $ and $\kernelFuncIntra$ with a kernel size of $1 \times \vert \SoAncGroup \vert$. This divides the domain of the kernel to two smaller domains: $\Point \times \{\mathrm{I}\}$ for $\kernelFuncInter$, and $\{\mathbf{0}\} \times G$ for $\kernelFuncIntra$, where $\mathrm{I}$ is the identity matrix, and $\mathbf{0}$ is a zero displacement vector. From here, we are ready to separate Eq.~\ref{equ:6d_conv_alt} into two convolutions:
\vspace{-2mm}
\begin{align}
    (\SoFeat \ast \kernelFuncInter)(\point, \soAnc) &= \sum_{\pointi^\prime \in \Point_{\soAnc}} \SoFeat(\point - \pointi^\prime, \soAnc) \kernelFuncInter(\soAnc \pointi^\prime, \mathrm{I})
    \label{equ:separate_inter} \\
    (\SoFeat \ast \kernelFuncIntra)(\point, \soAnc) &= \sum_{\soAncj \in \SoAncGroup} \SoFeat(\point, \soAncj^{-1}\soAnc) \kernelFuncIntra(\mathbf{0}, \soAncj)
  \label{equ:separate_intra}
\end{align}

\vspace{-3mm}
 We can see that $\kernelFuncInter$ is a kernel only varied by translation in the reference frame of $\soAnc$, and $\kernelFuncIntra$ is a kernel only varied by the rotation $\soAncj$. In the following text, we simplify them to $\kernelFuncInter(\soAnc \pointi^\prime)$ and $\kernelFuncIntra(\soAncj)$. The division here matches with the observation that the space \seThree{} can be factorized into two spaces $\mathbb{R}^{3}$ and \soThree{}. 
 Sequentially applying the two convolutions in Eq.~(\ref{equ:separate_inter}-\ref{equ:separate_intra}) approximates the 6D convolution in Eq.~\ref{equ:6d_conv_alt} (Fig.~\ref{fig:conv}(d)) while maintaining equivariance to \seThree{} (proofs provided in the supplementary materials). The working principle here is similar to that of the Inception module~\cite{szegedy2015going} and its follow-up works~\cite{chollet2017xception}, which have shown the promising property of separable convolutions in improving the network performance with reduced cost. We name the two consecutive convolutions as \textit{\convNameInter{}} and \textit{\convNameIntra{}}, respectively, as shown in Fig.~\ref{fig:conv}. We refer the combined convolutions as \textit{\convNameFull{} (\convNameShort{})}.
 Formally, the original 6D convolution is approximated by: $\SoFeat \ast \kernelFunc \approx (\SoFeat \ast \kernelFuncInter) \ast \kernelFuncIntra$.

A \seThree{} equivariant convolutional network can be realized by consecutive blocks of \convNameShort{}. 
The network consumes the input $\Point$ and produces a \textit{\seThree{} equivariant} feature for the point set. 
Since \convNameShort{} only takes functions defined on \seThree{} as input, for each point in the input point set, we set $\SoFeat(\point, \soAnc)=1$ for each $\soAnc \in G$. 
In the following sections, we discuss in details the form of kernel and how it can be localized for each convolution module.

\subsection{\ConvNameInter{}}
\label{sec:conv_inter}

Our \convNameInter{} layer aims at aggregating point spatial correlations locally under a rotation group element $\soAnc$. Let $\mathcal{N}_\point = \{ \pointi \in \Point\ \big\vert\ \lVert \point - \pointi \rVert \leqslant r \}$ be the set of neighbor points for point $\point$, with a radius $r$, the \ConvNameInter{} with localized kernel is:
\begin{equation}
  (\SoFeat \ast \kernelFuncInter)(\point, \soAnc) = \sum_{\pointi^\prime \in \mathcal{N}_{\soAnc\point}} \SoFeat(\point - \pointi^\prime, \soAnc) \kernelFuncInter(\soAnc \pointi^\prime),
  \label{equ:inter_conv}
\end{equation} 
\noindent where $\mathcal{N}_{\soAnc\point} = \{ \soAnc^{-1}(\point - \pointi) \vert \pointi \in \mathcal{N}_\point \}$ is the set of displacements to the neighbor points under a rotation $g$. $\kernelFuncInter$ is a kernel defined in a canonical neighbor space $\mathcal{B}^3_r$. Given that the convolution is computed as a spatial correlation under a rotation $\soAnc$, the form of the kernel can be naturally extended from any spatial kernel function. 
While our framework is general to support various spatial kernel definitions, we introduce two kernel formulations that are used in our implementation.
\vspace{-4mm}
\paragraph{Explicit kernels.} Given kernel size $\kernelSize$, we can define a set of kernel points $\{ \tilde{y}_k\}_\kernelSize$ evenly distributed in $\mathcal{B}_r^3$.
Each kernel point is associated with a kernel weight matrix $\kernelWeight_k \in \mathbb{R}^{D_{in} \times D_{out}}$, where $D_{in}$ and $D_{out}$ are the input and output channel, respectively.
Let $\interCorrelation(\cdot, \cdot)$ be the correlation function between two points, we have
\begin{equation}
  \kernelFuncInter(\pointi) = \sum_{k}^\kernelSize \interCorrelation(x_i,\tilde{y}_k) \kernelWeight_k.
  \label{equ:correlation}
\end{equation}
\noindent The correlation function $\interCorrelation(y, \tilde{y})$ can be either linear or Gaussian. For example, in the linear case described in \cite{thomas2018tensor}, $\interCorrelation(y, \tilde{y}) = \max(0, 1 - \frac{\lVert y - \tilde{y} \rVert}{\sigma})$, where $\sigma$ adjusts the bandwidth. 

\vspace{-4mm}
\paragraph{Implicit kernels.} The implicit formulation gives a function on point set that does not utilize parameterized kernels and is generally not considered a convolutional operation. Rather, spatial correlation is computed implicitly by concatenating the local frame coordinates of points to their corresponding features. In the \seThree{} equivariant extension, the local coordinates are also composed by a corresponding rotation $g$. The implicit filter for the input signal $\SoFeat$ is:
\vspace{-2mm}
\begin{equation} \label{equ:pnpp_conv}
\begin{split}
  \kernelFuncInter(\SoFeat (\point, \soAnc)) &= \sum_{\pointi \in \mathcal{N}_\point} \kernelFuncInter (\SoFeat(\pointi, \soAnc), \soAnc^{-1}\pointi) \\
  &= \sum_{\pointi \in \mathcal{N}_\point} \begin{bmatrix}
	\SoFeat(\pointi, \soAnc) \\
	\soAnc^{-1}\pointi
\end{bmatrix} \kernelWeight.
 \end{split}
\end{equation}

We believe that other choices of kernel functions can be naturally extended from these two examples. In our implementation of the network, we use the explicit kernel formulation in all convolutional layers. The last layer before the output block of our network filters point features globally and therefore utilizes the implicit formulation, as it scales better to process a larger set of point features.

\subsection{\ConvNameIntra{}}
\label{sec:conv_intra}

Given a discrete rotation group \SoAncGroup{}, the \convNameIntra{} computes \soThree{} correlation between the input signal and a kernel defined on the group domain.

We define a set of kernel rotation and their associate kernel weight matrices as $\mathcal{N}_\soAnc = \{ \soAnc_j \in \SoAncGroup \}_\kernelNum$ and $\{ \kernelWeight_j \in \mathbb{R}^{D_{in} \times D_{out}} \}_\kernelNum$, with the kernel size $\kernelNum = \lvert \mathcal{N}_\soAnc \rvert$. Thus the kernel is simply $\kernelFuncIntra(\soAncj) = \kernelWeight_j$. Our \convNameIntra{} layer aggregates information from neighboring rotation signals within the group, which is given by
\begin{equation}
  (\SoFeat \ast \kernelFuncIntra)(\point, \soAnc) = \sum_{\soAncj \in \mathcal{N}_\soAnc} \SoFeat(\point, \soAncj^{-1}\soAnc) \kernelFuncIntra(\soAncj).
\end{equation}
\vspace{-4mm}


In our implementation, the icosahedron group can be used as the discrete rotation group. The $K$ neighbor rotations are a subset of the group that is smallest in the corresponding angle. The computation can be accelerated by pre-computing the permutation index and only performing constant-time query with an index layer at run time.

\vspace{-3mm}
\paragraph{Complexity analysis.} 
As illustrated in Fig.~\ref{fig:conv}, by combining the two equivariance-preserving convolutions, we can achieve a similar effect with Eq.~\ref{equ:6d_conv} at a significantly lower computational cost. 
In particular, suppose we divide the original number of kernels $K$ into $K_p$ and $K_g$, the number of kernels in the point and group convolution; $C = C_i C_o$ where $C_{i}$ and $C_{o}$ are the number of input and output channels, $N = N_pN_a$ is the product of the number of points and the number of \soThree{} element in a rotation group. The naive 6D convolution requires a computational complexity of $O(K_pK_gCN)$. In contrast, the complexity of our approach is reduced to $O((K_p+K_g)CN)$, which could achieve orders-lower complexity compared to the naive solution especially with large $K_p$ and $K_g$.

\subsection{Shape Matching with Attention mechanism}
\label{sec:att}
	
In this section, we demonstrate how attention mechanism can be utilized to harness the power of equivariant feature. Given spatially pooled features that are equivariant to \soThree{}: $\SoFeat(\soAnc) : \SoAncGroup \rightarrow \mathbb{R}^{D}$, we define a rotation-based attention $A: \SoAncGroup \rightarrow \mathbb{R}$, $A(g) = \{a_g \vert \sum_{g \in \SoAncGroup} a_g = 1 \}$. 
\vspace{-4mm}
\paragraph{\soThree{} Detection.} Suppose a task requires the network to predict the pose $R \in \soThree{}$ of an input shape. When the attention weight is used as a probability score, the equivariant network turns the pose estimation task into a \soThree{} detection task, which is analogous to bounding box detection. Intuitively, each element from the discrete rotation group can be interpreted as an anchor. A two-branch network is used to classify whether the anchor is the "dominant rotation". Every anchor regresses a small rotational offset from its corresponding rotation. The multi-task loss for rotational regression is then given by:
\vspace{-1mm}
\begin{equation}
 	\mathcal{L}(a,u,R,R^u) = \mathcal{L}_{cls}(a,u) + \lambda[u=1]\mathcal{L}_{2}(R^u R^T)
 	\vspace{-1mm}
\end{equation}

\noindent where $a = \{a_g \vert g \in G\}$ are the predicted probabilities and $R$ are the predicted relative rotations. $u = \{u_g \vert g \in G\}$ is the ground-truth label with $u_g=1$ if $g$ is the nearest rotation to the target ground truth rotation $R_{GT}$. $R^u = \{R^u_g \vert \forall g \in G, R^u_g g = R_{GT} \} $ is the ground truth relative rotation.
\vspace{-4mm}

\paragraph{Group Attentive Pooling.}  Global pooling layers are integrated as part of the network for spatial reduction of the representation. As many common tasks, such as classification, benefit from rotation invariance of the learned feature, global pooling is utilized by most rotation-equivariant architectures to aggregate information into an invariant representation. 
To integrate attention mechanism with global pooling, we propose \textit{\poolingName{} (\poolingNameShort{})}, which is given by
\begin{equation}
  \SoFeat_{inv} = \frac{\sum_{\soAnc} \exp(a_{\soAnc} / T)\SoFeat_\SoAncGroup(\soAnc)}{\sum_{\soAnc}\exp(a_{\soAnc} / T)}, 
\end{equation}
\noindent where $\SoFeat_\SoAncGroup(\soAnc)$ and $a_{\soAnc}$ are the input rotation-equivariant feature and attention weight on rotation $\soAnc$. $T$ is a temperature score to control the sharpness of the function response.
As visualized in Fig.~\ref{fig:teaser}, the output feature is invariant given a rotated input point cloud.
The confidence weight $a$ can be learned by minimizing the loss
$
\mathcal{L} = \mathcal{L}_{task} + \lambda \mathcal{L}_{sa},
$
where $\mathcal{L}_{task}$ is a task-specific loss (e.g. cross-entropy loss for classification and triplet loss for correspondence matching); $\mathcal{L}_{sa}$ is a optional cross-entropy loss that encourages the network to learn the canonical axis from the candidate orientations when ground truth canonical pose is available for supervision. 


%
%

\subsection{Implementation Details}
\label{sec:implementation}
The core element of our network is the \convNameShort{} block as shown in Fig.~\ref{fig:conv}(d).
It consists of one \convNameInter{} and one \convNameIntra{} operator, with a batch normalization and a leaky ReLU activation inserted in between and after. 
We employ a 5-layer hierarchical convolutional network.
Each layer contains two \convNameShort{} blocks, with the first one being strided by a factor of $2$.
The network outputs spatially pooled features that are equivariant to the rotation group \SoAncGroup. It can be then pooled into an invariant feature through a \poolingNameShort{} layer. For the classification network, the feature is fed into a fully connected layer and a \textit{softmax} layer. For the task of metric learning, the feature is processed with an L2 normalization. We provide detailed network parameters and downsampling strategy in the supplemental materials.

\section{Experiments}
\label{sec:result}

We hypothesize that our approach is most suitable for tasks where the objects of interest are rotated arbitrarily. To this end, we evaluate our approach on two rotation-related datasets: the rotated Modelnet40 dataset~\cite{wu20153d} and the 3DMatch dataset~\cite{zeng20163dmatch}.
To ensure a fair comparison to previous works, in all experiments, we use the implementation provided by the authors or the reported statistics if no source code is available. We provide the training details of the experiments in the supplemental materials.


\subsection{Experiments on Rotated ModelNet40}
\label{sec:exp_cls}

\nothing{
\haiwei{Why do we evaluate on this dataset? Canonical poses are known, thus a showcase of scenario where rotation supervision improves the performance of equivariant network, whereas the non-equivariant ones would not be able to take advantage of this canonical pose information.}}

\begin{figure}[t]
 \begin{center}
  \centering
 \vspace{-25pt}
  \includegraphics[width=0.9\linewidth]{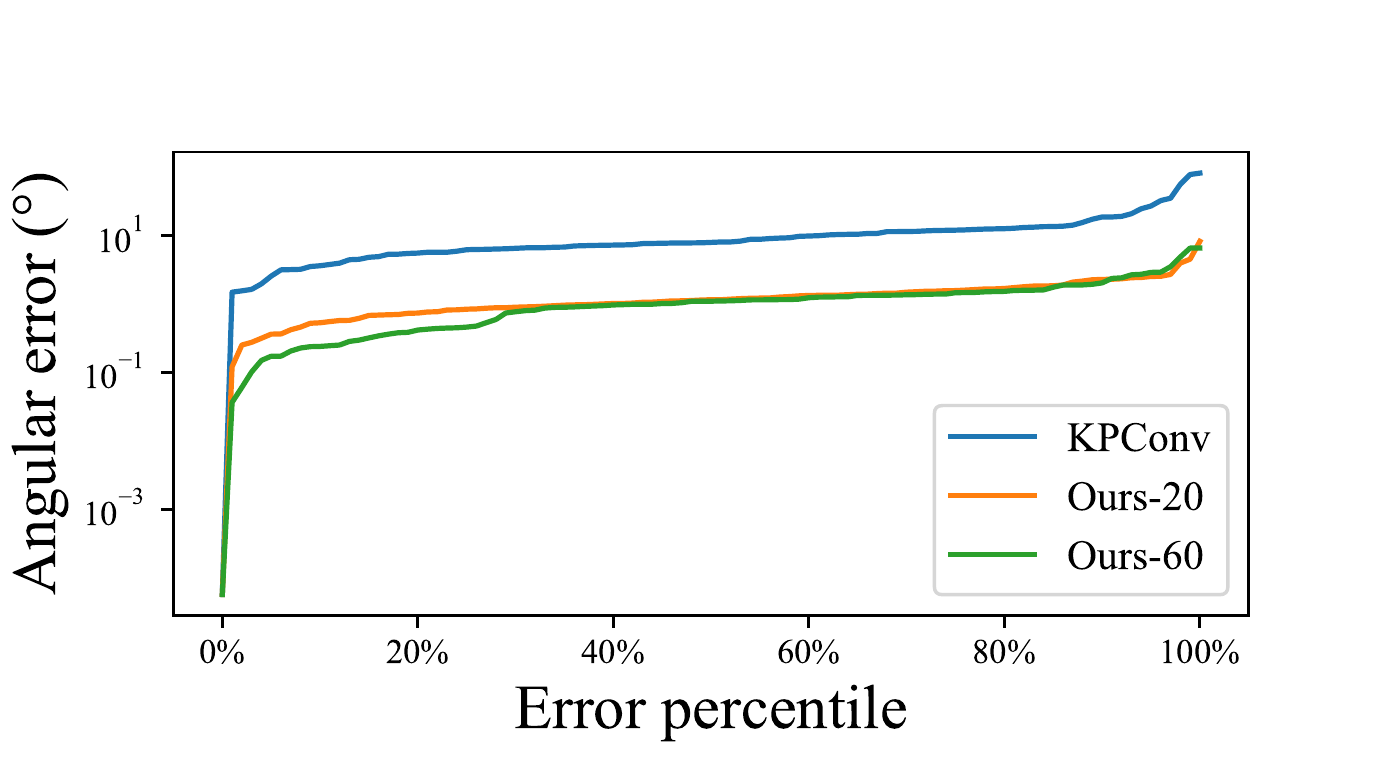}
 \end{center}
 \vspace{-20pt}
 \caption{Percentile of errors comparing KPConv~\cite{thomas2019kpconv} and two equivariant models (Ours-N) varied in number of \soThree{} elements.}
 \label{fig:pose}
 \vspace{-10pt}
\end{figure}

\paragraph{Dataset.} 
The official aligned Modelnet40 dataset provides a setting where canonical poses are known, and therefore it allows us to evaluate the effectiveness of pose supervision. We create the rotated ModelNet40 dataset based on the train/test split of the aligned ModelNet40 dataset~\cite{wu20153d}. 
We mainly focus on a more challenging ``rotated'' setting where each object is randomly rotated. 
For each object, we randomly subsample 1,024 points from the surface of the 3D object and perform random rotation augmentation before feeding it into the network. 
\vspace{-4mm}
\paragraph{Pose Estimation.} The pose estimation task predicts the rotation $\rotation \in \soThree$ that aligns a rotated input shape to its canonical pose. To avoid ambiguities indued by rotationally symmetric objects, we only use the airplane category from the dataset. We train the network with N=1252 airplane point clouds and test it with N=101 held-out point cloud, each augmented with random rotations. The evaluation compares equivariant models with KPConv~\cite{thomas2019kpconv}, a network that has similar kernel function to our implementation of point convolution, while not equivariant to 3D rotation. The equivariant models (Ours-N) are varied by the size of rotation group (N), similar to the setting in \cite{esteves2019equivariant}, and use the multitask detection loss described in Sec.~\ref{sec:att}. KPConv directly regresses the output rotation. Each model is trained for 80k iterations. The regressors in all models produce a rotation in the quaternion representation. 
We evaluate the performance by measuring angular errors between the predicted rotations and the ground-truth rotations. Tab.~\ref{tab:reg} shows the mean, median and max angular errors in each setting, and Fig.~\ref{fig:pose} plots the error percentile curves. As shown in the results, the equivariant networks significantly outperform the baseline network, with Ours-60 having the lowest errors. The equivariant networks also perform significantly more stable (max angular errors are kept within 9 degrees), while KPConv could produce unstable results for a certain inputs. This experiment showcases that a hierarchical rotation model can be much more effective in task that requires direct prediction of 3D rotation.


\begin{table}[t]
\small
\centering
\addtolength{\tabcolsep}{3pt}
\begin{tabular}{c|c|c|c}
\hline
  & Mean ($^{\circ}$) & Median ($^{\circ}$) & Max($^{\circ}$) \\ \hline
 KPConv~\cite{thomas2019kpconv} & 11.46 & 8.06 & 82.32\\
 Ours-20 & 1.36 & 1.16 & 8.30\\ 
 Ours-60 & \textbf{1.25} & \textbf{1.11} & \textbf{6.63} \\ \hline
\end{tabular}
\caption{Angular errors in point cloud pose estimation. 
}
\label{tab:reg}
\vspace{-1mm}
\end{table}

\begin{figure*}[]
 \begin{center}
  \includegraphics[width=0.85\linewidth]{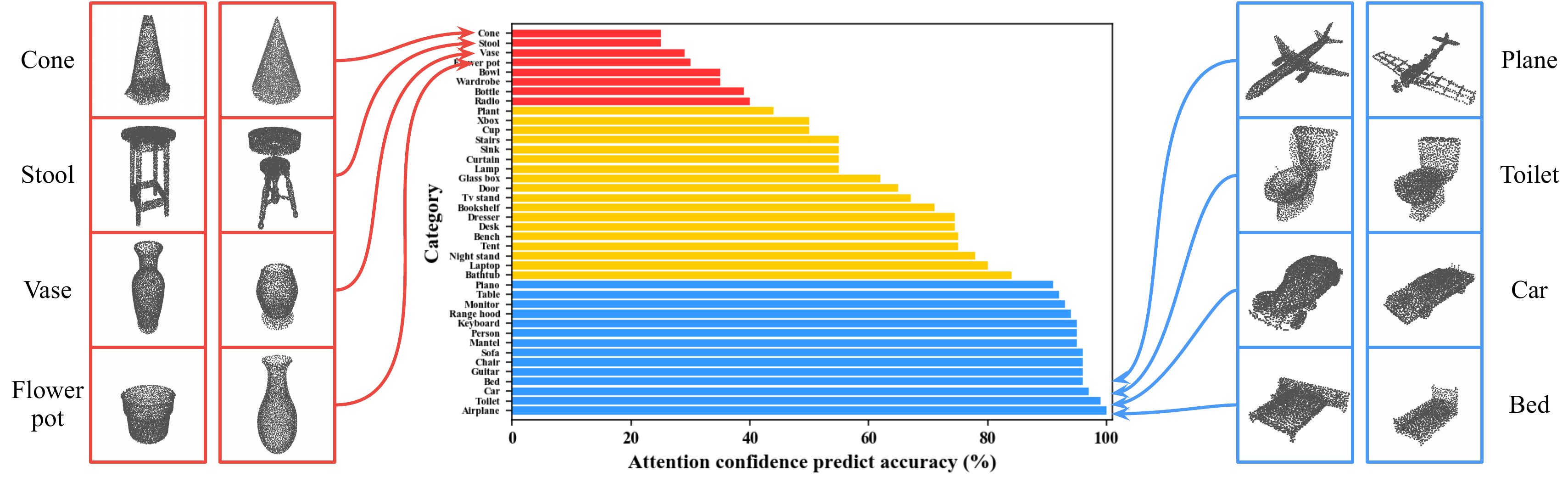}
 \end{center}
\vspace{-4mm}
\caption{Classification accuracy based on the attention confidence for each object category. The attention layer is trained on rotated dataset to learn a canonical orientation for the given object.}
\centering
 \label{fig:attconf}
\vspace{-4mm}
\end{figure*}


\begin{table}[]
\small
\centering
\addtolength{\tabcolsep}{-1pt}
\begin{tabular}{l|l|c|c}
\hline
\multicolumn{1}{c|}{Representation} & Methods & \multicolumn{1}{l|}{Acc (\%)} & \makecell{Retrieval \\ (mAP)} \\ \hline
\multirow{2}{*}{3D Surface}{\begin{tabular}[c]{@{}c@{}}  \end{tabular}} 
 & RotationNet~\cite{kanezaki2018rotationnet} & 80.0 & 74.2 \\
 & Sph. CNN~\cite{esteves2018learning} & 86.9 & - \\ 
 \hline
\multirow{6}{*}{Point cloud} 
 & QENet~\cite{zhao2019quaternion} & 74.4 & - \\
 & PointNet~\cite{qi2017pointnet} & 83.6 & - \\
 & PointNet++~\cite{qi2017pointnet++} & 85.0 & 70.3 \\
 & DGCNN~\cite{qi2017pointnet} & 81.1 & - \\
 & PointCNN~\cite{qi2017pointnet++} & 84.5 & - \\ 
 & KPConv~\cite{thomas2019kpconv} & 86.7 & 77.5 \\
 & Ours & \textbf{88.3} & \textbf{79.7} \\ \hline
\end{tabular}
\caption{Results on shape classification and retrieval on randomly rotated objects of ModelNet40. 
}
\label{tab:cls}
\vspace{-7mm}
\end{table}
\vspace{-4mm}
\paragraph{Classification and Retrieval.} The classification and retrieval tasks on Modelnet40 follow evaluation metric from~\cite{wu20153d}. In addition, our network is trained with \poolingNameShort{} and pose supervision introduced in Sec.~\ref{sec:att}. In Tab.~\ref{tab:cls}, we show the results comparing with the state-of-the-art methods in the setting where models are both trained and tested with rotation augmentation. We categorize the baseline approaches based on the input 3D representations: 3D surface and point cloud.

In the classification and retrieval task, our models also achieve the best performance, as shown in Tab.~\ref{tab:cls}. This indicates that our proposed framework can learn more effective and discriminative features even in the challenging cases that all the objects are randomly rotated.
\vspace{-3mm}
\paragraph{Ablation Analysis.} We further conduct an ablation study to validate the effectiveness of each algorithmic component. In particular, we experiment with five variants of our model by altering key designs in our network under the same architecture as shown in Tab.~\ref{tab:modelnet_abl}. By using the supervised attentive pooling, we can improve the classification accuracy with the same number of parameters compared to the max and mean pooling. However, the unsupervised attentive pooling does not outperform max pooling. This may be partly due to the difficulty of learning canonical pose in an unsupervised manner. In addition, only using point convolution will lead to a decline in performance, indicating the effectiveness of group convolution.
\begin{table}[]
\small
\addtolength{\tabcolsep}{2pt}
\centering
\begin{tabular}{c|c|c|c}
\hline
conv &  global pool & Loss & Acc (\%)   \\ \hline
Separable Conv & Attentive & $\mathcal{L}_{cls} + \mathcal{L}_{sa}$ & \textbf{88.3}\\ 
Separable Conv & Attentive & $\mathcal{L}_{cls}$  & 87.7\\ 
Separable Conv & Max & $\mathcal{L}_{cls}$  & 87.7\\ 
Separable Conv & Mean & $\mathcal{L}_{cls}$  &  87.4\\ 
Point Conv & Attentive & $\mathcal{L}_{cls} + \mathcal{L}_{sa}$  & 86.1\\ \hline
\end{tabular}
\caption{Results of ablation studies on ModelNet40 dataset. The \textit{conv} column denotes the configuration of convolution layers. The \textit{global pool} column denotes the type of global pooling method. Loss configuration follows notation from Sec.~\ref{sec:att}.
} 
\label{tab:modelnet_abl}
\vspace{-6mm}
\end{table}
%
%
%

\begin{table*}[h]
\small
\addtolength{\tabcolsep}{1.5pt}
\centering
\begin{tabular}{l|cccccc|cc|c}
\hline
        & SHOT\cite{tombari2010unique} & 3DMatch\cite{zeng20163dmatch} & CGF\cite{khoury2017learning}  & PPFNet\cite{deng2018ppfnet} & PPFF\cite{deng2018ppf} & 3DSNet\cite{gojcic2019perfect} & Li\cite{li2020end} & Li\cite{li2020end}$^\flat$ &   Ours \\ \hline
Kitchen & 74.3 & 58.3    & 60.3 & 89.7   & 78.7        & 97.5    & 92.1 & \textcolor{black}{\textbf{99.4}} & 99.0 \\  
Home 1  & 80.1 & 72.4    & 71.1 & 55.8   & 76.3        & 96.2            & 91.0 & \textcolor{black}{98.7} & \textbf{99.4} \\  
Home 2  & 70.7 & 61.5    & 56.7 & 59.1   & 61.5        & 93.2            & 85.6 & \textcolor{black}{94.7} & \textbf{96.2}\\  
Hotel 1 & 77.4 & 54.9    & 57.1 & 58.0   & 68.1        & 97.4            & 95.1 & \textcolor{black}{\textbf{99.6}} & \textbf{99.6}\\  
Hotel 2 & 72.1 & 48.1    & 53.8 & 57.7   & 71.2        & 92.8            & 91.3 & \textcolor{black}{\textbf{100.0}} & 97.1\\  
Hotel 3 & 85.2 & 61.1    & 83.3 & 61.1   & 94.4        & 98.2             & 96.3 & \textcolor{black}{\textbf{100.0}} & \textbf{100.0} \\ 
Study   & 64.0 & 51.7    & 37.7 & 53.4   & 62.0        & 95.0             & 91.8 & \textcolor{black}{95.5} & \textbf{96.2} \\ 
MIT Lab & 62.3 & 50.7    & 45.5 & 63.6   & 62.3        & \textbf{94.1}             & 84.4 & \textcolor{black}{92.2} & 93.5\\ \hline 
Average & 73.3 & 57.3    & 58.2 & 62.3   & 71.8        & 95.6             & 91.0 & \textcolor{black}{97.5} & \textbf{97.6} \\ \hline 
\end{tabular}
\caption{Comparisons of average recall of keypoint correspondences on 3DMatch. 
Li~\cite{li2020end}$^\flat$ denotes results tested with point normal information provided by the authors. All other results are tested on the official 3DMatch evaluation set without point normals.
}

\vspace{-5mm}
\label{tab:3dmatch}
\end{table*}
%
\vspace{-4mm}
\paragraph{How well does the attention layer learn?} It is possible that the performance of \poolingNameShort{} in distinguishing canonical poses could be compromised by the rotational symmetry of the object. If a shape is circularly symmetric, and the canonical poses prescribed by the rotational label is aligned with an axis of symmetry, the attention layer would naturally fail to provide a deterministic prediction. We summarize the classification accuracy based on the attention confidence for each category of ModelNet objects, as shown in Fig.~\ref{fig:attconf}. The results indeed support our intuition: the attention layer is ambiguous on objects with circular symmetry (e.g. cone and flower pot) and very confident on categories that have distinctive canonical orientation. On one hand, this shows that when the object of interest is asymmetric in rotation, the \poolingNameShort{} does help improve classification performance by establishing a local reference frame. On the other hand, the \poolingNameShort{} only fails at symmetric object that benefits relatively less from a equivariant representation. In the extreme case, the attention layer could be reduced to an average pooling.  
\vspace{-1mm}

\subsection{Shape Alignment on 3DMatch}
\label{sec:exp_3dm}

\nothing{
\haiwei{Why do we evaluate on this dataset? noisy, incomplete real scan, thus challenging. Fragments are under various rotations due to camera motion, only partially overlapped, thus would benefit from equivariant in both translation and rotation.}
}

\vspace{-1mm}
\paragraph{Dataset.} The 3DMatch dataset is a real-scan dataset consisting of 433 overlapping fragments from 8 indoor scenes for evaluation, and RGB-D data set from 62 indoor scenes for training. The pose of each fragment is determined by the camera angle during capturing, and two fragments at most overlap partially. Evaluating our model on this dataset is meaningful as shape registration in such setting would benefit from descriptors that are invariant to rigid camera motion. Each test fragment is a densely sampled point cloud with 150,000 to 600,000 points. To be consistent with our baselines, we use an evaluation metric based on the average recall of keypoints correspondence without performing RANSAC, following~\cite{deng2018ppfnet}. We also follow previous works~\cite{deng2018ppf,deng2018ppfnet,gojcic2019perfect} to set the matching threshold $\tau_1=0.1m$ and the inlier ratio $\tau_2=0.05$. 

\vspace{-4mm}
\paragraph{Comparison with baselines.} 
We designed a Siamese network for this task and trained our model with the batch-hard triplet loss proposed in \cite{gojcic2019perfect}. The input to the network is 1024-point patches extracted locally from keypoints in a fragment. The output is 64-dim invariant descriptors. Since a canonical ground truth pose is not known in this setting, the attentive pooling module in our model is trained in an unsupervised manner. 
Our results are shown in Tab.~\ref{tab:3dmatch}. 
To provide a comprehensive comparison, we select the state-of-the-art baselines using a variety of approaches: 1) convolutional network without rotational invariance, e.g.~\cite{zeng20163dmatch,deng2018ppfnet} 2) handcrafted invariant features w/ and w/o deep learning, e.g.~\cite{khoury2017learning,tombari2010unique,deng2018ppf}, 3) features learned from LRF aligned input \cite{gojcic2019perfect}, and 4) multi-view network \cite{li2020end}. We report the 64-dim results of \cite{gojcic2019perfect} to match the feature dimension of our model. Since the official 3DMatch test dataset does not contain point normal information, we report two results of \cite{li2020end}: a result of their model trained and tested without normal information (Li~\cite{li2020end} in Tab.~\ref{tab:3dmatch}) and one that is trained and tested with the authors' provided point normals (Li~\cite{li2020end}$^\flat$ in Tab.~\ref{tab:3dmatch}). We evaluate our model with the interest points provided by the authors of the dataset, which is consistent with the reported results of our baselines.    
Overall, our model outperforms all of the baselines in average recall, without the need to precompute an invariant representation or a local reference frame. Compare to some baselines (e.g. \cite{deng2018ppf, li2020end}) that requires dense point input, our model can learn discriminative features from very sparsely sampled sets of 1024 point. Our result is also better than the state-of-the-art method \cite{li2020end}, even without needing normal information as input. In the official setting where point normal information is not available, the performance of our model marks a great leap forward.

\begin{figure}[h]
 \begin{center}
  \centering
  \includegraphics[width=1.0\linewidth]{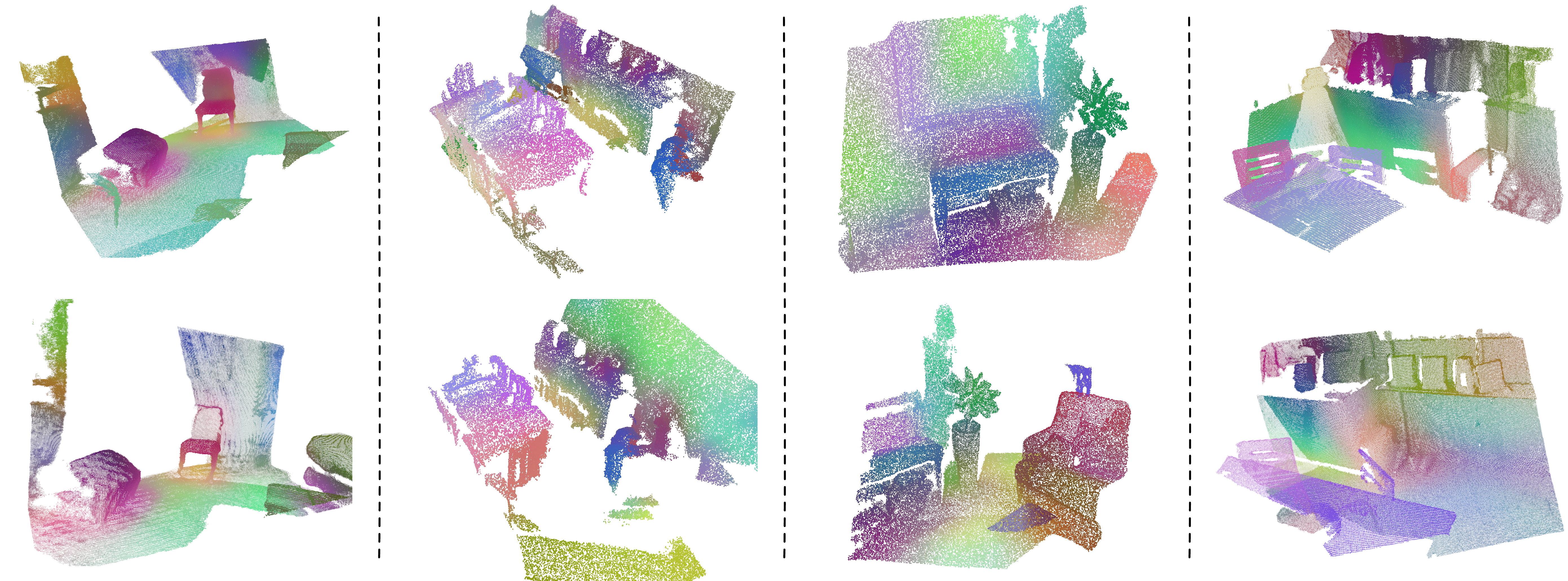}
 \end{center}
  \vspace{-15pt}
 \caption{T-SNE visualization of features learned by our network. Each column contains a pair of fragments from the same scene. Regions in correspondence are automatically labeled with similar features. }{}
 \vspace{-5mm}
 \label{fig:vis_3dmatch}
\end{figure}
\vspace{-5mm}
\paragraph{Qualitative analysis.} 
We provide a T-SNE visualization of the features learned by our network in Fig.~\ref{fig:vis_3dmatch}.
As different features are labeled with distinct colors, we can observe that the features learned by our network can robustly generate correct geometry correspondences even when the point cloud is incomplete, partially aligned,  or significantly rotated. For instance, in the third column, the bottom scene is only partially aligned with the top one and is viewed at an entirely different angle, our network can still reliably label the corresponded points with similar features. 

\vspace{-3mm}
\section{Conclusions and Discussions}
\label{sec:conclusion}

We have presented a novel framework that efficiently computes and leverages \seThree-equivariant features for point cloud analysis. First, we introduce a novel formulation named \seThree{} separable convolution that factorizes the naive \seThree{} convolution into two concatenated operators performed in two subspaces. Second, we propose the incorporation of attention mechanism that can appreciate and maintain the expressiveness of \seThree-equivariant features, which provides a novel way for 3D alignment tasks and can be used as a pooling layer that fuses the equivariant features into their more ready-to-use invariant counterparts. Such paradigm has led to leaps of performance in a variety of challenging tasks. Our approach is one of the earliest attempts of investigating \seThree-equivariant features for point cloud analysis. We believe there are still ample opportunities for more efficient methods and extension of the equivariant features to a broader range of applications.

\section*{Acknowledgements}
This research was sponsored by the Army Research Office and was accomplished under Cooperative Agreement Number W911NF-20-2-0053, and sponsored by the U.S. Army Research Laboratory (ARL) under contract number W911NF-14-D-0005, the CONIX Research Center, one of six centers in JUMP, a Semiconductor Research Corporation (SRC) program sponsored by DARPA and in part by the ONR YIP grant N00014-17-S-FO14. The views and conclusions contained in this document are those of the authors and should not be interpreted as representing the official policies, either expressed or implied, of the Army Research Office or the U.S. Government. The U.S. Government is authorized to reproduce and distribute reprints for Government purposes notwithstanding any copyright notation.

{\small
	\bibliographystyle{ieee_fullname}
	\bibliography{ms.bbl}

\begin{thebibliography}{10}\itemsep=-1pt

\bibitem{azulay2018deep}
Aharon Azulay and Yair Weiss.
\newblock Why do deep convolutional networks generalize so poorly to small
  image transformations?
\newblock {\em arXiv preprint arXiv:1805.12177}, 2018.

\bibitem{ben20183dmfv}
Yizhak Ben-Shabat, Michael Lindenbaum, and Anath Fischer.
\newblock 3dmfv: Three-dimensional point cloud classification in real-time
  using convolutional neural networks.
\newblock {\em IEEE Robotics and Automation Letters}, 3(4):3145--3152, 2018.

\bibitem{birdal2015point}
Tolga Birdal and Slobodan Ilic.
\newblock Point pair features based object detection and pose estimation
  revisited.
\newblock In {\em 2015 International Conference on 3D Vision}, pages 527--535.
  IEEE, 2015.

\bibitem{chollet2017xception}
Fran{\c{c}}ois Chollet.
\newblock Xception: Deep learning with depthwise separable convolutions.
\newblock In {\em Proceedings of the IEEE conference on computer vision and
  pattern recognition}, pages 1251--1258, 2017.

\bibitem{cohen2016group}
Taco Cohen and Max Welling.
\newblock Group equivariant convolutional networks.
\newblock In {\em International conference on machine learning}, pages
  2990--2999, 2016.

\bibitem{cohen2018spherical}
Taco~S. Cohen, Mario Geiger, Jonas Koehler, and Max Welling.
\newblock Spherical cnns.
\newblock 2018.
\newblock Proceedings of the 6th International Conference on Learning
  Representations (ICLR), 2018.

\bibitem{cohen2019gauge}
Taco~S Cohen, Maurice Weiler, Berkay Kicanaoglu, and Max Welling.
\newblock Gauge equivariant convolutional networks and the icosahedral cnn.
\newblock {\em arXiv preprint arXiv:1902.04615}, 2019.

\bibitem{cohen2016steerable}
Taco~S Cohen and Max Welling.
\newblock Steerable cnns.
\newblock {\em arXiv preprint arXiv:1612.08498}, 2016.

\bibitem{deng2018ppf}
Haowen Deng, Tolga Birdal, and Slobodan Ilic.
\newblock Ppf-foldnet: Unsupervised learning of rotation invariant 3d local
  descriptors.
\newblock In {\em Proceedings of the European Conference on Computer Vision
  (ECCV)}, pages 602--618, 2018.

\bibitem{deng2018ppfnet}
Haowen Deng, Tolga Birdal, and Slobodan Ilic.
\newblock Ppfnet: Global context aware local features for robust 3d point
  matching.
\newblock In {\em Proceedings of the IEEE Conference on Computer Vision and
  Pattern Recognition}, pages 195--205, 2018.

\bibitem{elbaz20173d}
Gil Elbaz, Tamar Avraham, and Anath Fischer.
\newblock 3d point cloud registration for localization using a deep neural
  network auto-encoder.
\newblock In {\em Proceedings of the IEEE Conference on Computer Vision and
  Pattern Recognition}, pages 4631--4640, 2017.

\bibitem{esteves2018learning}
Carlos Esteves, Christine Allen-Blanchette, Ameesh Makadia, and Kostas
  Daniilidis.
\newblock Learning so (3) equivariant representations with spherical cnns.
\newblock In {\em Proceedings of the European Conference on Computer Vision
  (ECCV)}, pages 52--68, 2018.

\bibitem{esteves2017polar}
Carlos Esteves, Christine Allen-Blanchette, Xiaowei Zhou, and Kostas
  Daniilidis.
\newblock Polar transformer networks.
\newblock {\em arXiv preprint arXiv:1709.01889}, 2017.

\bibitem{esteves2019equivariant}
Carlos Esteves, Yinshuang Xu, Christine Allen-Blanchette, and Kostas
  Daniilidis.
\newblock Equivariant multi-view networks.
\newblock In {\em Proceedings of the IEEE International Conference on Computer
  Vision}, pages 1568--1577, 2019.

\bibitem{frome2004recognizing}
Andrea Frome, Daniel Huber, Ravi Kolluri, Thomas B{\"u}low, and Jitendra Malik.
\newblock Recognizing objects in range data using regional point descriptors.
\newblock In {\em European conference on computer vision}, pages 224--237.
  Springer, 2004.

\bibitem{girshick2014rich}
Ross Girshick, Jeff Donahue, Trevor Darrell, and Jitendra Malik.
\newblock Rich feature hierarchies for accurate object detection and semantic
  segmentation.
\newblock In {\em Proceedings of the IEEE conference on computer vision and
  pattern recognition}, pages 580--587, 2014.

\bibitem{gojcic2019perfect}
Zan Gojcic, Caifa Zhou, Jan~D Wegner, and Andreas Wieser.
\newblock The perfect match: 3d point cloud matching with smoothed densities.
\newblock In {\em Proceedings of the IEEE Conference on Computer Vision and
  Pattern Recognition}, pages 5545--5554, 2019.

\bibitem{groh2018flex}
Fabian Groh, Patrick Wieschollek, and Hendrik~PA Lensch.
\newblock Flex-convolution.
\newblock In {\em Asian Conference on Computer Vision}, pages 105--122.
  Springer, 2018.

\bibitem{guo2016comprehensive}
Yulan Guo, Mohammed Bennamoun, Ferdous Sohel, Min Lu, Jianwei Wan, and
  Ngai~Ming Kwok.
\newblock A comprehensive performance evaluation of 3d local feature
  descriptors.
\newblock {\em International Journal of Computer Vision}, 116(1):66--89, 2016.

\bibitem{guo2013rotational}
Yulan Guo, Ferdous Sohel, Mohammed Bennamoun, Min Lu, and Jianwei Wan.
\newblock Rotational projection statistics for 3d local surface description and
  object recognition.
\newblock {\em International journal of computer vision}, 105(1):63--86, 2013.

\bibitem{guo2020deep}
Yulan Guo, Hanyun Wang, Qingyong Hu, Hao Liu, Li Liu, and Mohammed Bennamoun.
\newblock Deep learning for 3d point clouds: A survey.
\newblock {\em IEEE transactions on pattern analysis and machine intelligence},
  2020.

\bibitem{hua2018pointwise}
Binh-Son Hua, Minh-Khoi Tran, and Sai-Kit Yeung.
\newblock Pointwise convolutional neural networks.
\newblock In {\em Proceedings of the IEEE Conference on Computer Vision and
  Pattern Recognition}, pages 984--993, 2018.

\bibitem{huang2018learning}
Haibin Huang, Evangelos Kalogerakis, Siddhartha Chaudhuri, Duygu Ceylan,
  Vladimir~G Kim, and Ersin Yumer.
\newblock Learning local shape descriptors from part correspondences with
  multiview convolutional networks.
\newblock {\em ACM Transactions on Graphics (TOG)}, 37(1):6, 2018.

\bibitem{jiang2019spherical}
Chiyu Jiang, Jingwei Huang, Karthik Kashinath, Philip Marcus, Matthias
  Niessner, et~al.
\newblock Spherical cnns on unstructured grids.
\newblock {\em arXiv preprint arXiv:1901.02039}, 2019.

\bibitem{johnson1999using}
Andrew~E. Johnson and Martial Hebert.
\newblock Using spin images for efficient object recognition in cluttered 3d
  scenes.
\newblock {\em IEEE Transactions on pattern analysis and machine intelligence},
  21(5):433--449, 1999.

\bibitem{kanezaki2018rotationnet}
Asako Kanezaki, Yasuyuki Matsushita, and Yoshifumi Nishida.
\newblock Rotationnet: Joint object categorization and pose estimation using
  multiviews from unsupervised viewpoints.
\newblock In {\em Proceedings of the IEEE Conference on Computer Vision and
  Pattern Recognition}, pages 5010--5019, 2018.

\bibitem{khoury2017learning}
Marc Khoury, Qian-Yi Zhou, and Vladlen Koltun.
\newblock Learning compact geometric features.
\newblock In {\em Proceedings of the IEEE International Conference on Computer
  Vision}, pages 153--161, 2017.

\bibitem{lenssen2018group}
Jan~Eric Lenssen, Matthias Fey, and Pascal Libuschewski.
\newblock Group equivariant capsule networks.
\newblock In {\em Advances in Neural Information Processing Systems}, pages
  8844--8853, 2018.

\bibitem{li2019discrete}
Jiaxin Li, Yingcai Bi, and Gim~Hee Lee.
\newblock Discrete rotation equivariance for point cloud recognition.
\newblock {\em arXiv preprint arXiv:1904.00319}, 2019.

\bibitem{li2018so}
Jiaxin Li, Ben~M Chen, and Gim Hee~Lee.
\newblock So-net: Self-organizing network for point cloud analysis.
\newblock In {\em Proceedings of the IEEE conference on computer vision and
  pattern recognition}, pages 9397--9406, 2018.

\bibitem{li2018deep}
Junying Li, Zichen Yang, Haifeng Liu, and Deng Cai.
\newblock Deep rotation equivariant network.
\newblock {\em Neurocomputing}, 290:26--33, 2018.

\bibitem{li2020end}
Lei Li, Siyu Zhu, Hongbo Fu, Ping Tan, and Chiew-Lan Tai.
\newblock End-to-end learning local multi-view descriptors for 3d point clouds.
\newblock In {\em Proceedings of the IEEE/CVF Conference on Computer Vision and
  Pattern Recognition}, pages 1919--1928, 2020.

\bibitem{liu2019point2sequence}
Xinhai Liu, Zhizhong Han, Yu-Shen Liu, and Matthias Zwicker.
\newblock Point2sequence: Learning the shape representation of 3d point clouds
  with an attention-based sequence to sequence network.
\newblock In {\em Proceedings of the AAAI Conference on Artificial
  Intelligence}, volume~33, pages 8778--8785, 2019.

\bibitem{diego2017}
Diego Marcos, Michele Volpi, Nikos Komodakis, and Devis Tuia.
\newblock Rotation equivariant vector field networks.
\newblock In {\em Proceedings of the IEEE International Conference on Computer
  Vision}, pages 5048--5057, 2017.

\bibitem{maturana2015voxnet}
Daniel Maturana and Sebastian Scherer.
\newblock Voxnet: A 3d convolutional neural network for real-time object
  recognition.
\newblock In {\em 2015 IEEE/RSJ International Conference on Intelligent Robots
  and Systems (IROS)}, pages 922--928. IEEE, 2015.

\bibitem{qi2017pointnet}
Charles~R Qi, Hao Su, Kaichun Mo, and Leonidas~J Guibas.
\newblock Pointnet: Deep learning on point sets for 3d classification and
  segmentation.
\newblock In {\em Proceedings of the IEEE Conference on Computer Vision and
  Pattern Recognition}, pages 652--660, 2017.

\bibitem{qi2016volumetric}
Charles~R Qi, Hao Su, Matthias Nie{\ss}ner, Angela Dai, Mengyuan Yan, and
  Leonidas~J Guibas.
\newblock Volumetric and multi-view cnns for object classification on 3d data.
\newblock In {\em Proceedings of the IEEE conference on computer vision and
  pattern recognition}, pages 5648--5656, 2016.

\bibitem{qi2017pointnet++}
Charles~Ruizhongtai Qi, Li Yi, Hao Su, and Leonidas~J Guibas.
\newblock Pointnet++: Deep hierarchical feature learning on point sets in a
  metric space.
\newblock In {\em Advances in neural information processing systems}, pages
  5099--5108, 2017.

\bibitem{rusu2009fast}
Radu~Bogdan Rusu, Nico Blodow, and Michael Beetz.
\newblock Fast point feature histograms (fpfh) for 3d registration.
\newblock In {\em 2009 IEEE International Conference on Robotics and
  Automation}, pages 3212--3217. IEEE, 2009.

\bibitem{rusu2008aligning}
Radu~Bogdan Rusu, Nico Blodow, Zoltan~Csaba Marton, and Michael Beetz.
\newblock Aligning point cloud views using persistent feature histograms.
\newblock In {\em 2008 IEEE/RSJ International Conference on Intelligent Robots
  and Systems}, pages 3384--3391. IEEE, 2008.

\bibitem{spezialetti2019learning}
Riccardo Spezialetti, Samuele Salti, and Luigi~Di Stefano.
\newblock Learning an effective equivariant 3d descriptor without supervision.
\newblock In {\em Proceedings of the IEEE International Conference on Computer
  Vision}, pages 6401--6410, 2019.

\bibitem{szegedy2015going}
Christian Szegedy, Wei Liu, Yangqing Jia, Pierre Sermanet, Scott Reed, Dragomir
  Anguelov, Dumitru Erhan, Vincent Vanhoucke, and Andrew Rabinovich.
\newblock Going deeper with convolutions.
\newblock In {\em Proceedings of the IEEE conference on computer vision and
  pattern recognition}, pages 1--9, 2015.

\bibitem{thomas2019kpconv}
Hugues Thomas, Charles~R Qi, Jean-Emmanuel Deschaud, Beatriz Marcotegui,
  Fran{\c{c}}ois Goulette, and Leonidas~J Guibas.
\newblock Kpconv: Flexible and deformable convolution for point clouds.
\newblock {\em arXiv preprint arXiv:1904.08889}, 2019.

\bibitem{thomas2018tensor}
Nathaniel Thomas, Tess Smidt, Steven Kearnes, Lusann Yang, Li Li, Kai Kohlhoff,
  and Patrick Riley.
\newblock Tensor field networks: Rotation-and translation-equivariant neural
  networks for 3d point clouds.
\newblock {\em arXiv preprint arXiv:1802.08219}, 2018.

\bibitem{tombari2010unique}
Federico Tombari, Samuele Salti, and Luigi Di~Stefano.
\newblock Unique shape context for 3d data description.
\newblock In {\em Proceedings of the ACM workshop on 3D object retrieval},
  pages 57--62. ACM, 2010.

\bibitem{shot2010}
Federico Tombari, Samuele Salti, and Luigi Di~Stefano.
\newblock Unique signatures of histograms for local surface description.
\newblock In {\em European conference on computer vision}, pages 356--369.
  Springer, 2010.

\bibitem{weiler20183d}
Maurice Weiler, Mario Geiger, Max Welling, Wouter Boomsma, and Taco Cohen.
\newblock 3d steerable cnns: Learning rotationally equivariant features in
  volumetric data.
\newblock In {\em Advances in Neural Information Processing Systems}, pages
  10381--10392, 2018.

\bibitem{weiler2018learning}
Maurice Weiler, Fred~A Hamprecht, and Martin Storath.
\newblock Learning steerable filters for rotation equivariant cnns.
\newblock In {\em Proceedings of the IEEE Conference on Computer Vision and
  Pattern Recognition}, pages 849--858, 2018.

\bibitem{worrall2018cubenet}
Daniel Worrall and Gabriel Brostow.
\newblock Cubenet: Equivariance to 3d rotation and translation.
\newblock In {\em Proceedings of the European Conference on Computer Vision
  (ECCV)}, pages 567--584, 2018.

\bibitem{worrall2017harmonic}
Daniel~E Worrall, Stephan~J Garbin, Daniyar Turmukhambetov, and Gabriel~J
  Brostow.
\newblock Harmonic networks: Deep translation and rotation equivariance.
\newblock In {\em Proceedings of the IEEE Conference on Computer Vision and
  Pattern Recognition}, pages 5028--5037, 2017.

\bibitem{wu20153d}
Zhirong Wu, Shuran Song, Aditya Khosla, Fisher Yu, Linguang Zhang, Xiaoou Tang,
  and Jianxiong Xiao.
\newblock 3d shapenets: A deep representation for volumetric shapes.
\newblock In {\em Proceedings of the IEEE conference on computer vision and
  pattern recognition}, pages 1912--1920, 2015.

\bibitem{xu2018spidercnn}
Yifan Xu, Tianqi Fan, Mingye Xu, Long Zeng, and Yu Qiao.
\newblock Spidercnn: Deep learning on point sets with parameterized
  convolutional filters.
\newblock In {\em Proceedings of the European Conference on Computer Vision
  (ECCV)}, pages 87--102, 2018.

\bibitem{yang2017toldi}
Jiaqi Yang, Qian Zhang, Yang Xiao, and Zhiguo Cao.
\newblock Toldi: An effective and robust approach for 3d local shape
  description.
\newblock {\em Pattern Recognition}, 65:175--187, 2017.

\bibitem{zeng20163dmatch}
Andy Zeng, Shuran Song, Matthias Nie{\ss}ner, Matthew Fisher, Jianxiong Xiao,
  and Thomas Funkhouser.
\newblock 3dmatch: Learning local geometric descriptors from rgb-d
  reconstructions.
\newblock In {\em CVPR}, 2017.

\bibitem{zhao2019quaternion}
Yongheng Zhao, Tolga Birdal, Jan~Eric Lenssen, Emanuele Menegatti, Leonidas
  Guibas, and Federico Tombari.
\newblock Quaternion equivariant capsule networks for 3d point clouds.
\newblock {\em arXiv preprint arXiv:1912.12098}, 2019.

\end{thebibliography}
}

\ifthenelse{\equal{\final}{0}}
{
}
{}
\end{document}